\documentclass[conference]{IEEEtran}
\IEEEoverridecommandlockouts
\usepackage{cite}
\usepackage{amsmath,amssymb,amsfonts}
\usepackage{algorithm}
\usepackage{algorithmic}
\usepackage{graphicx}
\usepackage{textcomp}
\usepackage{xcolor}
\usepackage{stfloats}
\def\BibTeX{{\rm B\kern-.05em{\sc i\kern-.025em b}\kern-.08em
    T\kern-.1667em\lower.7ex\hbox{E}\kern-.125emX}}
\begin{document}

\title{PSO-PS:Parameter Synchronization with Particle Swarm Optimization for Distributed Training of Deep Neural Networks
\thanks{This work is supported in part by the National Key Research and Development Program of China under Contract 2017YFB1002201, in part by the National Natural Science Fund for Distinguished Young Scholar under Grant 61625204, and in part by the State Key Program of the National Science Foundation of China under Grant 61836006.
}
\thanks{Correspondence to Jiancheng Lv
}
}

\author{\IEEEauthorblockN{Qing Ye}
\IEEEauthorblockA{\textit{College of Computer Science} \\
  \textit{Sichuan University}\\
Chengdu, China}
\and
\IEEEauthorblockN{ Yuxuan Han}
\IEEEauthorblockA{\textit{College of Computer Science} \\
\textit{Sichuan University}\\
Chengdu, China}
\and
\IEEEauthorblockN{ Yanan Sun}
\IEEEauthorblockA{\textit{College of Computer Science} \\
\textit{Sichuan University}\\
Chengdu, China}
\and
\IEEEauthorblockN{ Jiancheng Lv}
\IEEEauthorblockA{\textit{College of Computer Science} \\
\textit{Sichuan University}\\
Chengdu, China}
\and

}

\maketitle
\begin{abstract}
Parameter updating is an important stage in parallelism-based distributed deep learning. Synchronous methods are widely used in distributed training the Deep Neural Networks (DNNs). To reduce the communication and synchronization overhead of synchronous methods, decreasing the synchronization frequency (e.g., every $n$ mini-batches) is a straightforward approach. However, it often suffers from poor convergence. In this paper, we propose a new algorithm of integrating Particle Swarm Optimization (PSO) into the distributed training process of DNNs to automatically compute new parameters. In the proposed algorithm, a computing work is encoded by a particle, the weights of DNNs and the training loss are modeled by the particle attributes. At each synchronization stage, the weights are updated by PSO from the sub weights gathered from all workers, instead of averaging the weights or the gradients. To verify the performance of the proposed algorithm, the experiments are performed on two commonly used image classification benchmarks: MNIST and CIFAR10, and compared with the peer competitors at multiple different synchronization configurations. The experimental results demonstrate the competitiveness of the proposed algorithm.  
\end{abstract}

\begin{IEEEkeywords}
Distributed training, PSO, SSGD
\end{IEEEkeywords}

\section{Introduction}
With the increasing growth of data volume and the complexity of neural networks, the efficient training of Deep Neural Networks (DNNs) has been becoming a challenging task for the community of machine learning. To address this issue, many distributed methods have been proposed to accelerate the training of DNNs, which can be generally divided into two different frameworks: data parallelism and model parallelism~\cite{Dean2012Large}. Particularly, the data parallelism refers to that the large dataset is divided into multiple different parts with small size, and the DNN is simultaneously trained on multiple different computational nodes with different dataset parts, while the model parallelism means that the DNN model is divided into different small models and each computational node performs a model on the entire dataset. In practice, the data parallelism has been attracted more attention because of its simplicity and easy to implement. In the data parallelism, the key problem is how to appropriately update the parameters (i.e., the gradients in terms of the weights of the DNNs) trained from different dataset parts during each iteration. The synchronous method is a classical approach to update the parameters,  where the server node distributes the workload to multiple nodes and then gathers all gradients in each iteration, which is also known as the Synchronous Stochastic Gradient Descent (SSGD)~\cite{CoatesDeep,Seide-1-bit}. The SSGD method is simple yet efficient. However, due to the conflicting between the synchronized settings and the possible variations that needs to compute a batch on different nodes at each iteration, all nodes have to wait for the slowest one to finish before entering the next iteration, which is ineffective. Furthermore, in each iteration, the parameter server needs to synchronize the multiple nodes, which will bring the communication and synchronization overhead, especially when the synchronization frequency is high or the number of computing nodes is large, which is inefficiency. 

To alleviate the burdens aforementioned, Povey \textit{et al.}~\cite{PoveyParallel} proposed a method by averaging the neural network parameters periodically (typically every one or two minutes). In addition, multiple state-of-the-art methods have also been developed by focusing on dramatically reducing the size of the exchanged gradients with slight convergence deterioration, such as the Asynchronous Parallel Stochastic Gradient Descent (APSGD)~\cite{Keuper2015Asynchronous}, the staleness-aware asyncSGD method~\cite{2015Staleness-aware}, the gradient sparsification method~\cite{ShiA}, the quantification method~\cite{HubaraQuantized,WenXu17Terngrad}, and the compression methods~\cite{LinDeep,2018arXiv180508768S}. Furthermore, some other state-of-the-art methods, concerning about conquering the communication challenge, have also been proposed, including the high-speed networks~\cite{2015arXiv151100175I}, such as the 10GbE and the InfiniBand are used to alleviate the communication cost. Recently, researchers have also proposed the approaches based on the communication efficient learning~\cite{TsuzukuVariance,2019arXiv190505957T} and the decentralized learning~\cite{Zhang2016HogWild,LianAsynchronous,2017arXiv170509056L}. Although these methods have experimentally demonstrated their promising performance in their respective papers, it is still hard to balance the communication overhead and convergence in the distributed training~\cite{PoveyParallel}.

Evolutionary computation is a class of nature-inspired computational paradigm and has shown their promising performance related to DNNs~\cite{Al-SahafA}, such as the design of neural architectures~\cite{Juang2004,SunA,Xie2017Genetic} because of their superiority performance for addressing complex optimization tasks. In this paper, we proposed to address the parameter updating problems by using the Particle Swarm Optimization (PSO) algorithm that is a widely used evolutionary computation method. To the best of our knowledge,  this is the first time to use the evolutionary computation methods to tackle the parameter updating in the distributed training of DNNs. Specifically, in the proposed algorithm, the training parameters of the distributed neural network are encoded into each particle of the PSO, and then the new parameters are calculated based on the particle updating mechanism, which is a completely different strategy from the existing methods that aggregate parameters or gradients at each synchronization. The goal of this paper is to discuss the potentiality of the proposed parameter synchronization method with PSO (PSO-PS) and treat it as an alternative approach to effectively solving the parameter updating problems in the distributed training. The contributions of the proposed PSO-PS algorithm are shown below:

\begin{itemize}
\item Propose an encoding strategy, which can incorporate the characteristics of PSO to well address the problems during distributed training the DNNs.
\item Propose an improved version of PSO to calculate the parameters because the neural network is sensitive to the variation of parameters. The improved PSO has an adjustable local and global search ability, which is very similar to the adjustable learning rate.
\item Experimentally investigate the proposed PSO-PS algorithm and demonstrate its competitive performance in distributed training the image classification benchmark datasets.
\end{itemize}

The remainder of this paper is organized as follows. The literature of the distributed deep learning (including the DNNs and the synchronous approaches with data parallelism for distributed training of DNNs) and PSO is reviewed in Section~\ref{section:two}. Section~\ref{section:three} illustrates the details of the proposed PSO-PS algorithm. The experiment design and the result analysis are documented in Section~\ref{section:four}. Finally, the conclusions and future work are drawn in Section~\ref{section:five}.

\section{Literature review}
\label{section:two}
\subsection{Deep Neural Networks (DNNs)}
DNNs are generally stacked with many hierarchical layers, and each layer represents a transformer function of the input. Mathematically, the DNNs can be formulated by Equation~(\ref{eq:dnn})

\begin{equation}\label{eq:dnn}
    a^{(l)}=f(w^{l},x^{l})
\end{equation}
where \(x^l\) and \(a^l\) denote the input and the output of the l-th layer, and the input of the current layer is the output of its previous layer (i.e., \(x^l\)=\(a^{l-1}\)). $f(\cdot)$  denotes the transformer function which consists of an operation (e.g., inner product or convolution) and an activation function (e.g., Sigmod). \(w^l\) denotes the trainable model parameters, which could be iteratively updated during the model training using mini-batch stochastic gradient descent (SGD) optimizers and the back propagation algorithm~\cite{Hecht2002Theory}.

Specifically, the SGD uses a small set of training samples (mini-batch), provides stable convergence at fair computational cost on a single node and can be formulated by Equation~(\ref{eq:sgd}):
\begin{equation}\label{eq:sgd}
   w^l=w^{l-1}-\eta \nabla w
\end{equation}

In Equation~(\ref{eq:sgd}), \(w^{l-1}\) represents the current iteration (e.g., weights), $\eta$ indicates the learning rate which is adjustable and $\nabla w$ is computed from a given loss-function (e.g., cross-entropy) over the forward results of the mini-batch. There are two ways to speed up the SGD: a) computing updates \(\nabla w\) with a faster way, and b) making a large batch size. Unfortunately, both methods will result in the unaffordable computation cost.
\subsection{Synchronous Approaches with Data Parallelism}
\label{section:PA}
Data parallelism is a classical framework to reduce the training time by keeping a copy of the entire DNN model on each worker (i.e., the computing node), processing different parts of the training data set on each worker. The data parallelism training approaches require well-designed methods to aggregate the results and synchronize the model parameters between each worker. Typically, there are two typical methods to achieve it: Parameter Averaging(PA) and Synchronous Stochastic Gradient Descent (SSGD).

\begin{figure}[h]
\centering
\includegraphics[width=7.4cm,height=4.5cm]{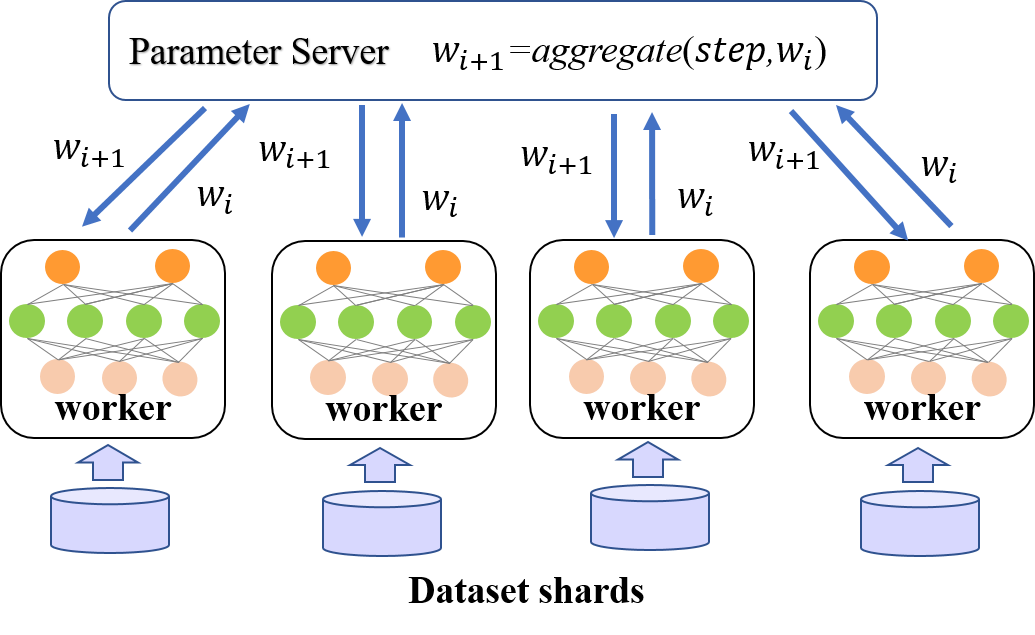}
\caption{Parameter averaging with data parallelism framework.}
\label{fig:dist}
\end{figure}

The PA is shown in Fig.~\ref{fig:dist}. Supposing there are $n$ workers, where $w$ indicates the parameters (weights, biases) of the DNN, the subscripts represent the iteration orders of the parameters, and the $step$ refers to the synchronization period. As can be seen from Fig.~\ref{fig:dist}, all the parameters are aggregated on the parameter server first, and then the new weights are produced. Next, the new weights are redistributed to the $n$ workers for the next iteration.

Specifically, the steps of the PA are provided below:
\begin{enumerate}
	\item Initialize the network parameters $w_i$ randomly based on the DNN configuration.
	\item Distribute a copy of the current parameters to each worker.
	\item Train each worker on a small part of the training dataset and get the sub weights.
	\item Collect and average all sub weights from each worker.
	\item Produce new weight by $\frac{1}{n}\sum w_{i}^{k}$ and distribute new weight~$w_{i+1}$ to all workers. $k$ represents the index of a worker, and $i$ represents the version of parameters.\label{sm:operation}
	\item Go to Step 3 until the termination is not satisfied.
\end{enumerate}

SSGD is similar to PA, the primary difference between SSGD and PA is that the SSGD transfers the gradient $\nabla w_{i}^{k}$ instead of the PA transferring parameters, i.e., in Step 5, the $w_{i+1}$ is updated by $w_i+\frac{1}{n} \alpha \sum \nabla w_{i}^{k} $ in the SSGD, and $\alpha$ is a scaling factor (analogous to a learning rate).

It is straightforward to prove that a restricted version of PA is mathematically identical to the SSGD during the training on a single machine when $step$ = 1. As can be seen from the details of the PA and the SSGD, the synchronous approaches are conceptually simple and are easy to synchronize the data after each iteration. However, in practice, the overhead of doing so is prohibitively high because the network communication and the synchronization cost may overwhelm the benefit obtained from the extra machines. Moreover, if the frequency of the synchronization is not reasonably specified, the local parameters in each worker may diverge too much, resulting in a poor model after the averaging.

\subsection{Particle Swarm Optimization(PSO)}
PSO is a population-based stochastic optimization approach, which was proposed by Eberhart and Kennedy in 1995~\cite{Eberhart2002A,Kennedy1995Particle}, by simulating the swarm behavior of birds and fish when they cooperatively search for the food. Especially, each member of the group (i.e., the particle) changes its search mode by learning its own experience and the experience of other members. The PSO algorithm initializes with a fix-sized set of particles that distribute across the solution space. On every step of the iterations, the locations of particles are passed to a shared function $F(\cdot)$  which calculates the fitness values. The particle with the highest value is marked, and the other particles should update its personal best record if the current value outperforms the values in its history. After that, all the particles are moved based on their previous locations according to the position updating mechanism formulated by Equations~(\ref{eq:pso1}) and~(\ref{eq:pso2}).
\begin{equation}\label{eq:pso1}
\begin{split}
    v_{id}^t=\begin{matrix} inertia \\ \overbrace{m*v_{id}^{t-1} }\end{matrix}+\begin{matrix} local\ search \\ \overbrace{c_1 r_1*(pBest_{id}-p_{id}^{t-1} )}\end{matrix} \\+ \begin{matrix}global\ search \\ \overbrace{c_2 r_2*(gBest_d-p_{id}^{t-1})}\end{matrix}
\end{split}
\end{equation}
\begin{equation}\label{eq:pso2}
    p_{id}^t=p_{id}^{t-1}+v_{id}^t
\end{equation}

In Equation~(\ref{eq:pso1}), $i$ represents the index of the particle and $t$ is the iteration counter. $m$ denotes non-negative inertia weight, $c_1$ and $c_2$ are the acceleration, constant $r$ is the random number between 0 and 1, $pBest_d$ denotes the fitness of local optimum particle in the $d^{th}$ dimension, and $gBest_d$ denotes the fitness of the global optimum particle in the $d^{th}$ dimension. The calculation of $v_{id}^{t}$ consists of three elements: inertia, local search and global search. Constants $m$, $c_1$, $r_1$ and $c_2$, $r_2$ are factors regulating the impact of three elements on the result, it means that three elements of current speed are adjustable in different application. The advantages of PSO are often viewed as its fast search speed and high efficiency, and the overall framework of the standard PSO is detailed as follows: 

\begin{enumerate}
\item Initialize particles:~population size~$n$, particle velocity~$v_i$, particle position~$p_i$, iteration counter~$t=1$; 
\label{ code:fram:init}

\item Evaluate the fitness of each particle:~$f_i$;
\label{code:fram:2}
\item Calculate the $gBest$ from the history of all particles by $f_i$;
\label{code:fram:3}
\item For each particle, calculate the best one $pBest_i$ from its memory by $f_i$;
\label{code:fram:4}
\item Updating velocity $v_i$ and position $p_i$ by Equations~(\ref{eq:pso1}) and ~(\ref{eq:pso2});
\label{code:fram:5}
\item Judge termination conditions. \textbf{If} $t > Max\_t$~(maximal generation number) \textbf{or} $f_i$ satisfy optimization target, go to step~7. \textbf{else}, repeat steps from~2 to~6. $t~=~t+1$;
\item Return the best position of the particle whose fitness value is $gBest$.
\end{enumerate}

In order to let PSO be more suitable for the distributed training of DNNs, we have made two improvements upon the standard PSO, and the improvement details are shown in Subsection~\ref{section:PSO-PS:partB}. 

\section{The proposed PSO-PS Algorithm}
\label{section:three}
PSO serves as a tool for optimization when different solutions to a problem are given and an evaluation metric is defined. The similarity between neural network training and swarm optimization lies in the fact that we don't know where the global optimization is, but we can keep approaching it, which inspires us to use PSO to facilitate the distributed training of DNNs. The purpose of aggregating all parameters in the distributed training is to make use of the training results of all workers. As mentioned in the Subsection~\ref{section:PA}, if the synchronization executes at each iteration, the distributed training is equal to a single machine training, this is also the most time-consuming phase. As the synchronization period grows, the communication overhead decreases, but averaging parameters may lead to a poor model. To this end, using PSO to update parameters instead of average can leave the best parameters and optimize other parameters at the same time, it is beneficial that all workers approach the optimal solution with lower communication cost.

\subsection{Encoding Strategy}
Developing an encoding strategy between the PSO and the distributed training of DNNs is the first step to use PSO.

As have mentioned above, the problem investigated in this paper is about the parameter updating for the data parallelism by using PSO. To achieve this, firstly, the whole input dataset is divided into equally sized chunks, where $n$ indicates the number of workers in the cluster. In the PSO, the size of the particle swarm is specified as the number of the workers $n$. Secondly, a different subset of data is fed into different workers, and each worker performs its forward pass and backward pass individually. Because the parameters of the DNN trained on each work will change as the training progresses. The parameters $w$ trained on the worker are modelled as the position of a particle. Clearly, the solution space is defined as all the possible weight combinations for the network needed to be trained. The fitness function is defined as the loss function of the network, so the loss on the current mini-batch corresponds to the fitness of a particle. The lower the fitness is, the closer the particle is to the global optimum, which is consistent with the goal of training the DNNs. Thirdly, when the calculation on several iterations is done, gradients are gathered up, averaged and synchronized across the cluster. At this step, PSO calculates new weights according to Equations~(\ref{eq:pso1}) and~(\ref{eq:pso2}), instead of averaging and synchronous operation. Finally, new weights are redistributed to each worker to continue training. Table~\ref{tab:modeling strategy} shows the detail of the encoded parameters of the proposed PSO-PS.

\begin{table}[tp]
	\centering
	\caption{Detail of the encoding strategy}
	\label{tab:modeling strategy}
	\begin{tabular}{|c|c|}
		\hline
		\textbf{PSO}& \textbf{Distributed deep learning}\\
		\hline
		particle population $N$ & distributed scale $n$\\
		\hline
		$p_i$&$w_i$\\
		\hline
		 $fitness$ & $training$ $loss$\\
		\hline
		$gBest$ & $min$ \{$loss_i$\}, $i \in n$\\
		\hline
		$pBest_i$ & $min$ \{${loss_i}^t$\}, $t$:from $0$ to current iteration\\
		
		\hline
	\end{tabular}
\end{table}

\subsection{Parameter Synchronization with PSO (PSO-PS)}
\label{section:PSO-PS:partB}
In this section, we will introduce the algorithm flow of PSO-PS in detail. In the standard PSO, many hyper-parameters such as $c_1$, $c_2$, and $m$, are constants. It means that global and local search capabilities remain constant throughout the calculation. At the early stage, the fixed parameters benefit convergence, while the value of the loss function will hover around the minimum value during the later phase, and it is difficult to reach the global optimal value all the time. To avoid the algorithm crossing the global optimum of DNN and slower convergence, two improvements have been proposed in this paper to enhance PSO-PS.

a) To speed up the convergence of PSO, the linear decline of weight proposed by Shi \textit{et al.}~\cite{Shi1998A} will be applied to the proposed PSO-PS algorithm. In Equation~(\ref{eq:pso_line_dec}), $m_{max}$ and $m_{min}$ are hyper-parameters. $t$ represents the current iteration, $t_{max}$ represent the maximum number of iterations. At the initial iteration, because the $t$ is small, the inertia weight $m$ is relatively large. It is advantageous to locate the approximate position of the optimal solution quickly. With the accumulation of iteration times, the value of $m$ becomes small, and particles slow down, which benefits local search.
\begin{equation}\label{eq:pso_line_dec}
m=m_{max}-t*\frac{m_{max}-m_{min}}{t_{max}}\qquad
\end{equation}

b) To decrease the randomness of PSO-PS, we introduce an additional variable $\lambda$ (range 1 to $epoch\_size$) to adjust the offset of parameters on every mini-batch (i.e., weaken random variability of weights in DNN training). As the epoch number increases, the random variable factor decays, and the local and global search abilities are weakened. Finally, each particle updates its speed and location (weights) according to Equations~(\ref{eq:dnn_pso1}) and~(\ref{eq:dnn_pso2}) :

\begin{equation}\label{eq:dnn_pso1}
v_{id}^t = m *v_{id}^{t-1}+ \frac{c_1 r_1}{\lambda} (pBest_{id}-w_{id}^{t-1})+\frac{c_2 r_2}{\lambda}(gBest_d-w_{id}^{t-1})
\end{equation}
\begin{equation}\label{eq:dnn_pso2}
w_{id}^t=w_{id}^{t-1}+v_{id}^t
\end{equation}
\begin{algorithm}
    \caption{Parameter Synchronization With PSO}
    \label{alg:PSO-PS}
    \begin{algorithmic}[1]
    \STATE Initialize raw weights ($w$)of the DNN according to configuration of DNN.
    \label{step:cluster:init}
    \STATE Initialize particles population by Algorithm~\ref{alg:PSO-PS-PI}
    \label{step:dnn_pso:init}
    \STATE $iteration\_size = \frac{training-size}{batch-size}$, $Max\_t \leftarrow epoch$-$size* iteration\_size$
    \label{step:dnn_pso:max_t}
    \STATE $t = 1$
    \WHILE {$t < Max\_t$ }
    \IF{ $t \% step == 0$}
    \STATE calculate $gBest$ by Algorithm~\ref{alg:gBest}
    \label{step:dnn_pso:gbest}
        \FOR{each particle $i \in N$}
        \IF{$loss_{i}^{t} < pBest\_loss_i$}
        \STATE $pBest_i$ = $w_{i}^{t}$;~$pBest\_loss_i$ = $loss_{i}^{t}$
        \label{step:dnn_pso:pbest}
        \ENDIF
        \STATE calculate the new weights: $w_{i}^{t+1}$ according to Equations~(\ref{eq:dnn_pso1}),~(\ref{eq:dnn_pso2})
        \label{step:dnn_pso:update}
        \ENDFOR
        \STATE distribute new weights $w_{i}^{t+1}$ to workers
        \label{step:dnn_pso:new weights}
    \ELSE
    \STATE update $w$ by SGD
    \label{step:pso_dnn:sgd}
    \ENDIF
    \STATE $t=t+1$
    \ENDWHILE 
    \STATE return $gBest$
    \end{algorithmic}
\end{algorithm}
    
Algorithm~\ref{alg:PSO-PS} shows the procedure of the PSO-PS algorithm. Step~\ref{step:cluster:init} completes the initialization of clusters with the same initialized parameters. Every work holds a replica of the entire DNN model and a different part of the dataset. Step~\ref{step:dnn_pso:init} completes the initialization of particles population. At step~\ref{step:dnn_pso:max_t}, $Max\_t$ represents the maximum iterations, $training\_size$ indicates the size of training dataset. To decrease the training time and utilize the gradients and the SGD optimizer, which has become the mainly used method in optimizing neural networks, and control the frequency of synchronization. $step$ is applied to make the optimizer switch between PSO and SGD. On every $step$ iteration, PSO-PS carries out once. Step~\ref{step:dnn_pso:pbest} complete the local search. All the workers execute key steps from~\ref{step:dnn_pso:gbest} to~\ref{step:dnn_pso:update} of PSO-PS to update new weights. $i$ represents the index of the worker, $t$ represent the version of parameters. On other iterations, each worker just applies the gradients on its mini-batch at step~\ref{step:pso_dnn:sgd}, ignoring other workers. New weights are redistributed to workers at step~\ref{step:dnn_pso:new weights}. When $step$ = 1, the algorithm falls back to be a full synchronous method, and when $step$ = $t_{max}$, it can be viewed as a cluster of workers all running SGD separately on their subset of the data, $step$ can be exploited to adjust the communication cost of the cluster.

\subsection{Initialization of PSO-PS}
The typical PSO initialization method is based on random initialization with the whole search space. In the PSO-PS, the population is initialized by the parameters of a worker. The initialization is detailed in the Algorithm~\ref{alg:PSO-PS-PI}. At step~\ref{step:pso-ps:size}, the population size is set as the size of the cluster scale. For example, If the cluster has $16$ workers, this means that the particle population size is $16$. At step~\ref{step:pso-ps:Initialization}, each particle information $p_i$ will completely clone parameters $w_i$ of DNN on a worker, including the structure and the value. The dimension of particle and parameters are the same. At the initialization stage, every worker hosts a replica of the DNNs model, so all particles have the same information.

\begin{algorithm}
    \caption{Particle  Initialization}
    \label{alg:PSO-PS-PI}
    \begin{algorithmic}[1]
    \STATE Particle population$\leftarrow$ distributed scale $n$
    \label{step:pso-ps:size}
    \STATE $i = 1$ //index of worker
    \FOR{each worker $i \in n$}
    \STATE  $p_i \leftarrow w_i$ 
    \label{step:pso-ps:Initialization}
    \ENDFOR
    \RETURN Initialized population $P$
    \end{algorithmic}
\end{algorithm}

\subsection{Global Search}
Since each worker holds a different subset of the dataset, parameters $w_i$ of each work will be changing as the training continues, the parameter of each network is constantly approaching the global optimal. It means the particle population is moving to the global optimal solution in the searching space. It is generally known that the loss function is used to measure the performance of the current model in the process of DNN training. In the PSO-PS, we need a function to validate the fitness of a particle. According to the encoding strategy in Table~\ref{tab:modeling strategy}, the fitness function is modelled by the loss function, the training loss of the DNN represents the fitness of a particle. Accordingly, the global search of PSO means to find the minimum training loss of all workers. The calculation of $gBest$ is detailed in Algorithm~\ref{alg:gBest}. There is no server in the cluster, all processes are equal. The communication operation adopts the AllReduce Algorithm~\cite{zhao-Kylix}, which is an efficient way to communicate and implemented by Pytorch. Step~\ref{step:AllReduce} gathers all particles and fitness by AllReduce. Step~\ref{step:minimum} gets the index of minimum loss by a function $agrmin$. Step~\ref{step:gBest} gets the best parameters in the cluster, which is the best particle at the current iteration.
\begin{algorithm}
    \caption{Global Search}
    \label{alg:gBest}
    \begin{algorithmic}[1]
    \STATE Initialize two lists, the size of a list is $n$: $parameter\_list$ and $loss\_list$, 
    \STATE Gather parameters and loss values of all workers by AllReduce.
    \label{step:AllReduce}
    \FOR{each worker $i \in n$}
    \STATE $parameter\_list[i] = w_i$ 
    \STATE $loss\_list[i] = loss_i$ 
    \label{step:gBest:init}
    \ENDFOR
    \STATE Get the index of the worker which has the minimum training loss: $gBest\_index$ = agrmin($loss\_list$) 
    \label{step:minimum}
    \RETURN $parameter\_list[gBest\_index]$
    \label{step:gBest}
    \end{algorithmic}
\end{algorithm}

\section{Experiment Design}
\label{section:four}
In this section, we provide the details of the benchmark dataset chosen, the classical neural network model and the parameter settings used in the experiments for investigating the performance of PSO-PS.

All experiments are performed on a single Tesla V100 machine with 4 GPUs. Multi-processes are used to simulate the multi-nodes in the distributed environment, i.e., each process runs as a single node. We validate the effectiveness of the proposed algorithm with two image classification benchmark datasets: MNIST~\cite{726791} and CIFAR10~\cite{CIFAR-10}. The MNIST dataset is a hand-written digit recognition dataset to classify the numeral numbers between 0 and 9, including a training set of 60,000 examples, and a test set of 10,000 examples. The CIFAR10 dataset comprises of a total of 60,000 RGB images of size 32$\ast$32 pixels partitioned into the training set (50,000 images) divided into five training batches and the test set (10,000 images) containing exactly 1000 randomly selected images from each class. Each image belongs to one of the 10 classes, with 6000 images per class.
\begin{table}[bt]
\centering
\caption{The parameter settings of PSO-PS}
\label{tab:parameters setting}
\begin{tabular}{|c|c|}
\hline
$m_{max}$&0.9\\
\hline
$m_{min}$&0.3\\
\hline
$c_1$& 0.2\\
\hline
$c_2$ &0.9\\
\hline
$r_1$,$r_2$ &    $random(0,1)$\\ 
\hline
$\lambda$ & $epoch$-$size$\\
\hline
\end{tabular}
\end{table}

We use two classical convolutional neural network(CNN) models: Let-Net~\cite{Lecun-Gradient-based} and ResNet~\cite{He2016Deep} to verify the effectiveness of the proposed PSO-PS algorithm. The Let-Net is often viewed as the first successful CNN model, which was designed to identify the hand-written digits in the MNIST dataset and it has 0.665 million parameters. The ResNet was proposed to address the depth issue by training a slightly different inter-layer interaction: instead of composing layers, every convolutional module would add its input to the output. Residuals are implemented as ``shortcut” identity connections in the ResNet network. It is possible to train networks with depths from 50 to 152 layers with ResNet, further increasing the quality of the results by allowing higher-level features to be learned. The ResNet model used in the experiments has 11.1 million parameters. The DNN model is trained By Adam optimizer~\cite{Adam} with momentum, and the loss function is specified to the cross-entropy. The dataset is partitioned according to the number of processes $n$ and each process holds one piece of the dataset. The parameters setting of the PSO-PS used for the experiment are detailed in Table~\ref{tab:parameters setting}.
\begin{table}[b]
\centering
\caption{The classification accuracy (in \%) of proposed PSO-PS and peer competitor SSGD on the MNIST dataset}
\label{tab:mnist result}
\begin{tabular}{|c|c|c|c|}
\hline
Algorithm & Avg(\%) & Max(\%) &min(\%) \\
\hline
SSGD $n$=4 & 98.25& 99.0&98.0\\
\hline
PSO-PS,$n$=4 &\textbf{99.0} &99.0    &99.0\\
\hline
SSGD,$n$=8 & 98.0 & 98.0 & 98.0\\
\hline
PSO-PS,$n$=8 &    \textbf{98.0} &    98.0&    98.0\\
\hline
SSGD,$n$=16 & 97.0 &    97.0 &    97.0\\
\hline
PSO-PS,$n$=16&\textbf{97.93}&    98.0 & 97.0\\
\hline 
\end{tabular}
\end{table}
\subsection{Evaluation on MNIST}
\label{exp:MNIST}

\begin{figure*}[htp]
	\centerline{\includegraphics[width=20cm]{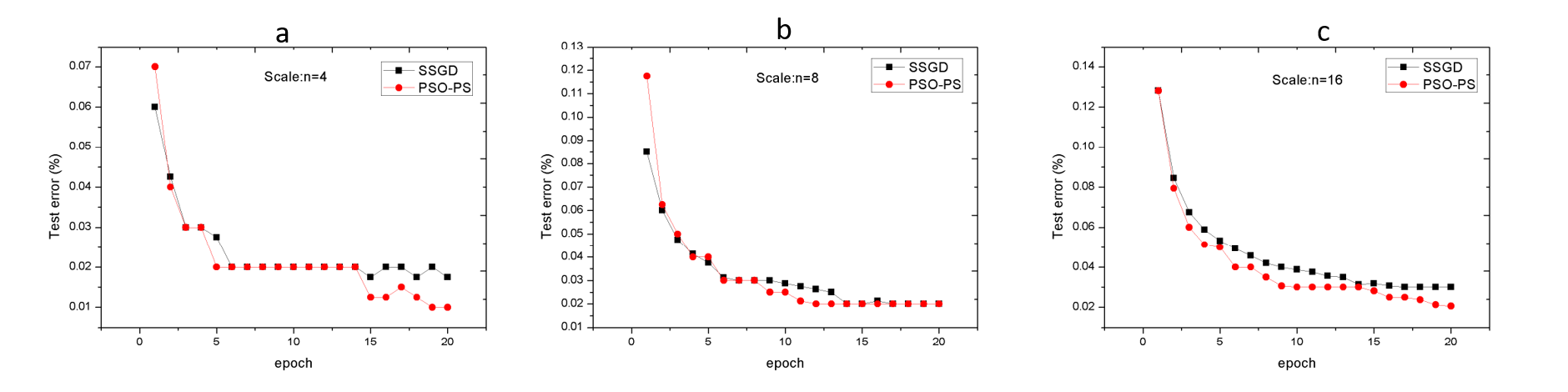}}
	\caption{Test error of SSGD and PSO-PS on MNIST with different scales.}
	\label{fig:mnist}
\end{figure*}

\begin{figure*}[htbp]
	\centerline{\includegraphics[width=20cm]{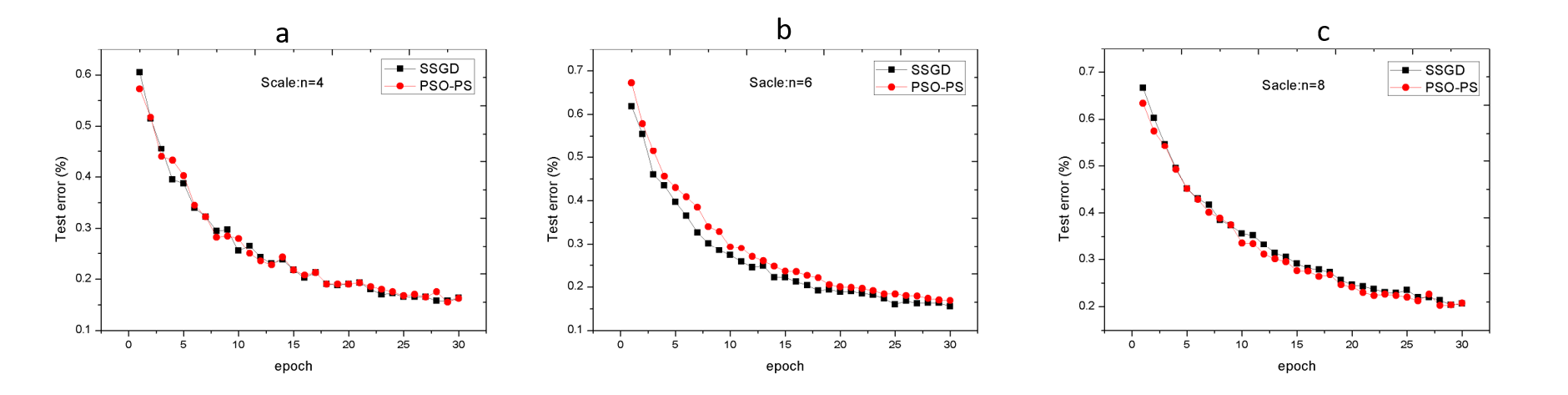}}
	\caption{Test error of SSGD and PSO-PS on CIFAR10 with different scales.}
	\label{fig:Res}
\end{figure*}

To the best of our knowledge, it is the first time that the PSO is used to optimize parameter synchronization in distributed training, so the most important goal in this paper is to verify whether the PSO-PS has the same convergence effect as that of SSGD or not, which is achieved by conducting the experiments for verifying the accuracy.

We train the Let-Net using both SSGD and PSO-PS on MNIST, where SSGD is used to provide the target model accuracy baseline for PSO-PS since it guarantees zero gradient staleness and achieves the best model accuracy. The variable $n$ represents the distributed scale. As suggested in~\cite{PoveyParallel}, we set $step$ = 10, $batch$-$size$ = 256, and $epoch$-$size$ = 25. Especially, the scale of the distributed cluster (i.e.,) can be changed according to specific computing resources. In the lab environment, the scale size is specified as 4, 8, and 16.

The experimental results are shown in Table~\ref{tab:mnist result}, where the first column indicates the SSGD and PSO-PS with different distributed cluster scale $n$, the second column indicates the average accuracy of the cluster, and the best accuracy and the worst accuracy of all workers are in the third and fourth columns, respectively. As can be seen from Table~\ref{tab:mnist result}, the accuracy of PSO-PS is better than SSGD, when scale size $n$ = 4 and 16. Meanwhile, Fig.~\ref{fig:mnist} shows the records of test error during the training process on MNIST using PSO-PS and SSGD with different scales. As shown in Figs.~\ref{fig:mnist}(a) and~\ref{fig:mnist}(b), the accuracy of PSO-PS is comparable to SSGD on MNIST dataset. Moreover, when the distributed scale becomes large, the accuracy of the PSO-PS outperforms SSGD in the training process as shown in Fig.~\ref{fig:mnist}(c), which is consistent with the characteristics of PSO, i.e., increasing the population properly is beneficial to the convergence of PSO, but the accuracy of PSO-PS do not improve with the increase of $n$. For example, scale 8 means the cluster extends 2 nodes compared to scale 6, but the accuracy of PSO-PS and the SSGD is equal. We think the main reason is that PSO is a random optimization approach with instability. In general, the PSO-PS and SSGD have the same convergence while the performance of the PSO-PS slightly outperforms the SSGD. 
\subsection{Evaluation on CIFAR10}
\label{exp:cifar}
In this part, we will investigate the performance of PSO-PS on a more complex network and benchmark dataset. Specifically, the experiments are designed on the training of ResNet on CIFAR10. In order to obtain the results in an acceptable training time, we set $step$ = 10, $batch$-$size$ = 256, $epoch$-$size$ = 30, and make a comparison between PSO-PS and SSGD on the optimization process with different distributed scales. In particular, the scale size is specified as 4, 6, and 8 in the lab environment. 

The convergence curves are shown in Fig.~\ref{fig:Res} with different scales. In Fig.~\ref{fig:Res}(b), although PSO-PS shows a slightly worse result than the baseline, PSO-PS converges faster than SSGD during the first several epochs, and runs the similar performance to the baseline when the scales are 4 and 8 (as shown in Figs.~\ref{fig:Res}(a) and~\ref{fig:Res}(c)). The main reason behind this is that the local and global search capability of PSO-PS can accelerate the training early, but this acceleration effect can not be constantly kept, especially in the later epochs when the network parameters are getting closer to their global minimum thus optimization becomes harder. In addition, the dimension of the ResNet parameters is up to millions, thus it's difficult to optimize well. To summarize, the convergence of a complex DNN  trained by PSO-PS is competitive to those of the baseline.

According to the above experiments, we can summarize the following points:
\begin{itemize}
\item When the parameter scale of the neural network is small, the performance of PSO-PS slightly outperforms the SSGD. 
\item As the parameter scale of the neural network increases, it means the data dimension that particles need to optimize increases, sometimes up to millions, it makes optimization more difficult. PSO-PS can accelerate the training at the early several epochs faster than SSGD, but this acceleration is unstable.
\item In general, with the increase of the cluster size, the advantages of PSO-PS are more significant, but it is also affected by the size of neural network parameters and the partition of the dataset.
\end{itemize}

\section{Conclusions and future work}
\label{section:five}
The goal of this paper is to design a novel parameter synchronization algorithm to accelerate the convergence of the neural networks in the distributed training. To achieve this, the PSO is introduced into distributed training (PSO-PS), which could be an alternative method to the parameter synchronization in distributed training. Specifically, we first proposed an encoding strategy to applies the PSO to DNN distributed training, then developed an improved PSO, which is more suitable for DNN distributed training. Finally, several experiments had been conducted to investigate the performance of PSO-PS. PSO-PS can work well with different scales of neural networks. Particularly, when the DNN scale is small, its convergence rate is faster than SSGD. Despite the interesting results we have obtained, the proposed PSO-PS also suffers from several drawbacks, for example, the acceleration effect of PSO-PS is not always stable, and the scale of the cluster simulating the population size of the particle is small relatively. These potential improvements are left for future work.

\bibliographystyle{IEEEtran}
\bibliography{IEEEabrv,PSO-PS}
\end{document}